\documentclass[10pt,twocolumn,letterpaper]{article}

\usepackage{cvpr}
\usepackage{times}
\usepackage{epsfig}
\usepackage{graphicx}
\usepackage{makecell}
\usepackage{amsmath}
\usepackage{amssymb}

\usepackage{paralist}

\cvprfinalcopy % *** Uncomment this line for the final submission

\usepackage[usenames,dvipsnames,svgnames,table]{xcolor}

\ifcvprfinal\usepackage[breaklinks=true,bookmarks=false]{hyperref}
\else\usepackage[pagebackref=true,breaklinks=true,letterpaper=true,colorlinks,bookmarks=false]{hyperref}\fi

\def\cvprPaperID{1457}

\newcommand{\st}{\textrm{s.t.,}}

\newtheorem{theorem}{Theorem}

\graphicspath{{images/}}
\DeclareGraphicsExtensions{.png}

\ifcvprfinal\fi % set for final copy
\begin{document}

\title{A Faster Method for Tracking and Scoring Videos Corresponding to
  Sentences}

\author{Haonan Yu\\
Purdue University\\
{\tt\small yu239@purdue.edu}
\and
Daniel P. Barrett\\
Purdue University\\
{\tt\small dpbarret@purdue.edu}
\and
Jeffrey Mark Siskind\\
Purdue University\\
{\tt\small qobi@purdue.edu}}

\maketitle

\begin{abstract}
Prior work presented the sentence tracker, a method for scoring how well a
sentence describes a video clip or alternatively how well a video clip depicts
a sentence.
We present an improved method for optimizing the same cost function employed by
this prior work, reducing the space complexity from exponential in the sentence
length to polynomial, as well as producing a qualitatively identical result in
time polynomial in the sentence length instead of exponential.
Since this new method is plug-compatible with the prior method, it can be used
for the same applications: video retrieval with sentential queries, generating
sentential descriptions of video clips, and focusing the attention of a tracker
with a sentence, while allowing these applications to scale with significantly
larger numbers of object detections, word meanings modeled with HMMs with
significantly larger numbers of states, and significantly longer sentences,
with no appreciable degradation in quality of results.
\end{abstract}

\section{Introduction}

We present an algorithmic improvement to the method of Siddharth
\etal\ \cite{siddharth2014}.
This prior work presented the \textsl{sentence tracker}, a method for scoring
how well a sentence describes a video clip or alternatively how well a video
clip depicts a sentence.
This method operates by applying an object detector to each frame of the video
clip to detect and localize instances of the nouns in the sentence and stringing
these detections together into tracks that satisfy the conditions of the
sentence.
To compensate for false negatives in the object-detection process, the
detectors are biased to overgenerate; the tracker must then select a single
best detection for each noun in the sentence in each frame of the video clip
that, when assembled into a collection of tracks, best depicts the sentence.
This prior work presented both a cost function for doing this as well as an
algorithm for finding the optimum of this cost function.
While this algorithm is guaranteed to find the global optimum of this cost
function, the space and time needed is exponential in the number of nouns in the
sentence, \ie\ the number of simultaneous objects to track.
Here, we present an improved method for optimizing the same cost function.
We prove that a relaxed form of the cost function has the same global optimum
as the original cost function.
We empirically demonstrate that local search on the relaxed cost function
finds a local optimum that is qualitatively close to the global optimum.
Moreover, this local search method takes space that is only linear in the
number of nouns in the sentence.
Each iteration takes time that is also only linear in the number of nouns in
the sentence.
In practice, the search process converges quickly.

This result is important because the sentence tracker, as a scoring function,
supports three novel applications \cite{barbu2014,siddharth2014}: the ability
to focus the attention of a tracker with a sentence that describes which
actions and associated participants to track in a video clip that depicts
multiple such, the ability to generate rich sentential descriptions of video
clips with nouns, adjectives, verbs, prepositions, and adverbs, and the ability
to search for video clips, in a large video database, that satisfy such rich
sentential queries.
Since the method presented here optimizes the same cost function, it yields
essentially identical scoring results, allowing it to apply in a
plug-compatible fashion, unchanged, to all three of these applications,
allowing them to scale to significantly larger problems.

\section{Background}

The sentence tracker is based on the \textsl{event tracker} \cite{Barbu2012b}
which is, in turn, based on detection-based tracking
\cite{avidan2004,han2004,Wolf1989,wu2007detection}.
The general idea is to bias an object detector to overgenerate, producing~$J^t$
detections, denoted $b^t_1,\ldots,b^t_{J^t}$, for each frame
$t\in\{1,\ldots,T\}$ in the video clip.
Each detection~$b$ has a score~$f(b)$, higher scores indicating greater
confidence.
Moreover, there is a measure~$g(b',b)$ of temporal coherence between detections
in adjacent frames.
If~$b'$ is a detection in frame $t-1$ and $b$~is a detection in frame~$t$,
higher values of $g(b',b)$ indicate that the position of~$b$ relative to~$b'$
is consistent with observed motion in the video, between frames~$t-1$ and~$t$,
such as may be computed with optical flow.
Detection-based tracking seeks to find a detection index~$j^t$ for each
frame~$t$ such that the track $\mathbf{j}=j^1,\ldots,j^T$, composed of the
selected detections $b^t_{j^t}$, maximizes both the overall detection score and
the overall temporal coherence.
One way that this can be done is by adopting the following cost function:
\begin{equation}
  \max_{\mathbf{j}}
  \left(\sum_{t=1}^Tf(b^t_{j^t})\right)+
  \left(\sum_{t=2}^Tg(b^{t-1}_{j^{t-1}},b^t_{j^t})\right)
  \label{eq:tracker}
\end{equation}
The advantage of this cost function is that a global optimum can be found in
polynomial time with the Viterbi algorithm \cite{viterbi1967,viterbi1971}.
This is done with dynamic programming \cite{bellman1957} on a lattice whose
columns are frames and whose rows are detections.
%
% yu239: we have T in space because we need to keep track of the optimal path
%
The overall time complexity of this approach is $O(TJ^2)$ and the overall space
complexity is $O(TJ)$, where~$J$ is the maximal number of detections per frame.

The event tracker \cite{Barbu2012b} extends this approach by adding a term to
the cost function that measures the degree to which the track depicts an event.
Events are modeled as HMMs
\cite{baum1966,chang2002extract,SiskindM96,Starner98,Wang2009,YamotoOI92}.
Eq.~\ref{eq:tracker} is analogous to the MAP estimate for an HMM over a
track~$\hat{\mathbf{j}}$:
\begin{equation}
  \max_{\mathbf{k}}
  \left(\sum_{t=1}^Th(k^t,b^t_{\hat{\jmath}^t})\right)+
  \left(\sum_{t=2}^Ta(k^{t-1},k^t)\right)
  \label{eq:map}
\end{equation}
where $a(k',k)$ denotes the transition probability from state~$k'$ to~$k$,
$k^t$ denotes the state in frame~$t$, and~$h(k,b)$ denotes the probability of
generating a detection~$b$ in state~$k$.
A global MAP estimate for the optimal state sequence
$\mathbf{k}=k^1,\ldots,k^T$ can also be found with the Viterbi algorithm, on a
lattice whose columns are frames and whose rows are states, in time $O(TK^2)$
and space $O(TK)$, where~$K$ is the number of states.

The event tracker operates by jointly optimizing the objectives of
Eqs.~\ref{eq:tracker} and~\ref{eq:map}:
\begin{equation}
  \max_{\mathbf{j},\mathbf{k}}
  \begin{array}[t]{l}
    \left(\displaystyle\sum_{t=1}^Tf(b^t_{j^t})\right)+
    \left(\displaystyle\sum_{t=2}^Tg(b^{t-1}_{j^{t-1}},b^t_{j^t})\right)\\
    {}+\left(\displaystyle\sum_{t=1}^Th(k^t,b^t_{j^t})\right)+
    \left(\displaystyle\sum_{t=2}^Ta(k^{t-1},k^t)\right)
  \end{array}
  \label{eq:event}
\end{equation}
The global optimum of this cost function can also be found with the Viterbi
algorithm using a lattice whose columns are frames and whose rows are pairs of
detections and states in time $O(T(JK)^2)$ and space $O(TJK)$.
This finds a track that not only has high scoring detections and is temporally
coherent but also exhibits the spatiotemporal characteristics of the event as
modeled by the HMM.\@

The sentence tracker forms a factorial \cite{brand1997,zhong2001} event tracker
with factors for multiple tracks to represent multiple event participants and
multiple HMMs to represent the meanings of multiple words in a sentence to
mutually constrain the overall spatiotemporal characteristics of a collection
of event participants to satisfy the semantics of a sentence.
Different pairs of participants are constrained by different words in the
sentence.
For example, the sentence \emph{The person to the left of the chair carried the
  backpack away from the traffic cone towards the stool} mutually constrains
the spatiotemporal characteristics of five participants: a \textbf{person}, a
\textbf{chair}, a \textbf{backpack}, a \textbf{traffic cone}, and a
\textbf{stool} with the pairwise constituent spatiotemporal relations
\begin{quote}
  \begin{tabular}{l}
    $\textsc{toTheLeftOf}(\textbf{person},\textbf{chair})$\\
    $\textsc{carried}(\textbf{person},\textbf{backpack})$\\
    $\textsc{awayFrom}(\textbf{backpack},\textbf{traffic cone})$\\
    $\textsc{towards}(\textbf{backpack},\textbf{stool})$\\
  \end{tabular}
\end{quote}
The sentence tracker operates by optimizing the following cost function:
\begin{equation}
  \max_{\mathbf{J},\mathbf{K}}
  \begin{array}[t]{l}
    \displaystyle
    \left[\sum_{l=1}^L
      \left(\sum_{t=1}^Tf(b^t_{j^t_l})\right)+
      \left(\sum_{t=2}^Tg(b^{t-1}_{j^{t-1}_l},b^t_{j^t_l})\right)\right]\\
    \displaystyle
        {}+\left[\begin{array}{l}\displaystyle\sum_{w=1}^W
          \begin{array}[t]{@{}l@{}}
            \left(\displaystyle\sum_{t=1}^T
            h_w
            (k^t_w,b^t_{j^t_{\theta^1_w}},\ldots,b^t_{j^t_{\theta^{I_w}_w}})\right)\\
            {}+\left(\displaystyle\sum_{t=2}^Ta_w(k^{t-1}_w,k^t_w)\right)
          \end{array}\end{array}\right]
  \end{array}
  \label{eq:sentence}
\end{equation}
where there are~$L$ event participants constrained by~$W$ content words,
$\mathbf{J}$~denotes the track collection $\mathbf{j}_1,\ldots,\mathbf{j}_L$,
and $\mathbf{K}$~denotes the state-sequence collection
$\mathbf{k}_1,\ldots,\mathbf{k}_W$.
In the above, the HMM output model $h(k,b_{j_{\theta^1}},\ldots,b_{j_{\theta^I}})$
is generalized to take more than one detection as input.
This is to model spatiotemporal relations of arity~$I$ between multiple
participants (typically two).
The \textsl{linking function}~$\theta$ specifies which participants apply in
which order and is derived by syntactic and semantic analysis of the sentence.

While a global optimum to this cost function can still be found with the
Viterbi algorithm, using a lattice whose columns are frames and whose rows are
tuples containing~$L$ detections and~$W$ states, the time complexity of such
is $O(T(J^LK^W)^2)$ and the space complexity is $O(TJ^LK^W)$.
Thus both the space and time complexity is exponential in~$L$ and~$W$, values
which increase linearly with the number of nouns in the sentence.

\section{Overview of the new method}

\begin{figure*}
  \begin{equation}
    \max_{\mathbf{P}}
    \max_{\mathbf{Q}}
    \begin{array}[t]{l}
      \left[\begin{array}{l}\displaystyle\sum_{l=1}^L
          \begin{array}[t]{l}
            \left(\displaystyle\sum_{t=1}^T
            \displaystyle\sum_{j=1}^{J^t} p^{t,l}_j f(b^t_j)\right)\\
                             {}+\left(\displaystyle\sum_{t=2}^T
                             \displaystyle\sum_{j'=1}^{J^{t-1}}
                             \displaystyle\sum_{j=1}^{J^t}
                             p^{t-1,l}_{j'} p^{t,l}_j
                             g(b^{t-1}_{j'},b^t_j)\right)\\
        \end{array}\end{array}\right]\\
           {}+\left[\begin{array}{l}\displaystyle\sum_{w=1}^W
               \begin{array}[t]{l}
                 \left(\displaystyle\sum_{t=1}^T
                 \displaystyle\sum_{j_1=1}^{J^t}\cdots
                 \displaystyle\sum_{j_{I_w}=1}^{J^t}
                 \displaystyle\sum_{k=1}^K
                 q^{t,w}_k p^{t,{\theta^1_w}}\cdots p^{t,{\theta^{I_w}_w}}_j
                 h_w(k,b^t_{j_1},\ldots,b^t_{j_{I_w}})\right)\\
                 {}+\left(\displaystyle\sum_{t=2}^T
                 \displaystyle\sum_{k'=1}^{K_w}
                 \displaystyle\sum_{k=1}^{K_w}
                 q^{t-1,w}_{k'} q^{t,w}_k a_w(k',k)\right)
             \end{array}\end{array}\right]
    \end{array}
    \label{eq:sentence-relax}
  \end{equation}
\end{figure*}

We reformulate Eqs.~\ref{eq:tracker}, \ref{eq:map}, and~\ref{eq:event} using
indicator variables instead of indices.
Instead of using~$j^t$ to indicate the index of a detection~$b^t_{j^t}$ in
frame~$t$, we use~$p^t_j$ as an indicator variable, which is zero for all
indices~$j$, except the index of the selected detection, for which it is one.
This allows Eq.~\ref{eq:tracker} to be reformulated as:
\begin{equation}
  \max_{\mathbf{p}}
  \begin{array}[t]{l}
    \left(\displaystyle\sum_{t=1}^T
    \displaystyle\sum_{j=1}^{J^t} p^t_j f(b^t_j)\right)\\
                     {}+\left(\displaystyle\sum_{t=2}^T
                     \displaystyle\sum_{j'=1}^{J^{t-1}}
                     \displaystyle\sum_{j=1}^{J^t}
                     p^{t-1}_{j'} p^t_j g(b^{t-1}_{j'},b^t_j)\right)
  \end{array}
  \label{eq:tracker-relax}
\end{equation}
Similarly, instead of using~$k^t$ to indicate the state in frame~$t$, we
use~$q^t_k$ as an indicator variable, which is zero for all states~$k$, except
that which is selected in frame~$t$, for which it is one.
This allows Eq.~\ref{eq:map} to be reformulated as:
\begin{equation}
  \max_{\mathbf{q}}
  \begin{array}[t]{l}
    \left(\displaystyle\sum_{t=1}^T
    \displaystyle\sum_{k=1}^K q^t_k h(k,b^t_{\hat{\jmath}^t}\right)\\
                     {}+\left(\displaystyle\sum_{t=2}^T
                     \displaystyle\sum_{k'=1}^K
                     \displaystyle\sum_{k=1}^K q^{t-1}_{k'} q^t_k a(k',k)\right)
  \end{array}
  \label{eq:map-relax}
\end{equation}
Combining the two allows Eq.~\ref{eq:event} to be reformulated as:
\begin{equation}
  \max_{\mathbf{p}}
  \max_{\mathbf{q}}
  \begin{array}[t]{l}
    \left(\displaystyle\sum_{t=1}^T
    \displaystyle\sum_{j=1}^{J^t} p^t_j f(b^t_j)\right)\\
    {}+\left(\displaystyle\sum_{t=2}^T
    \displaystyle\sum_{j'=1}^{J^{t-1}}
    \displaystyle\sum_{j=1}^{J^t} p^{t-1}_{j'} p^t_j g(b^{t-1}_{j'},b^t_j)\right)\\
    {}+\left(\displaystyle\sum_{t=1}^T
    \displaystyle\sum_{j=1}^{J^t}
    \displaystyle\sum_{k=1}^K q^t_k p^t_j h(k,b^t_j)\right)\\
    {}+\left(\displaystyle\sum_{t=2}^T
    \displaystyle\sum_{k'=1}^K
    \displaystyle\sum_{k=1}^K q^{t-1}_{k'} q^t_k a(k',k)\right)
  \end{array}
  \label{eq:event-relax}
\end{equation}
In the above, $\mathbf{p}$~and~$\mathbf{q}$ denote the collections of the
indicator variables~$p^t_j$ and~$q^t_k$ respectively.
One can view the indicator variables~$p^t_j$ and~$q^t_k$ to be constrained to
be in $\{0,1\}$ and to satisfy the sum-to-one constraints $\sum_jp^t_j=1$ and
$\sum_kq^t_k=1$.
Under these constraints, it is obvious that the formulations underlying
Eqs.~\ref{eq:tracker} and~\ref{eq:tracker-relax}, as well as
Eqs.~\ref{eq:map} and~\ref{eq:map-relax}, are identical.
One can further apply a similar transformation to Eq.~\ref{eq:sentence} to get
Eq.~\ref{eq:sentence-relax}, where the indicator variables~$p^{t,l}_j$ are
further indexed by the track~$l$ and the indicator variables~$q^{t,w}_k$ are
further indexed by the word~$w$.
$\mathbf{P}$~and~$\mathbf{Q}$ denote the collections of all indicator
variables~$p^{t,l}_j$ and~$q^{t,w}_k$ respectively.
Similarly, it is again clear that the formulations underlying
Eqs.~\ref{eq:sentence} and~\ref{eq:sentence-relax} are identical.
Note that because $h(k,b_{j_{\theta^1}},\ldots,b_{j_{\theta^I}})$ is generalized to
take more than one detection as input, the objective in
Eq.~\ref{eq:sentence-relax} becomes a multivariate polynomial of degree~$I+1$,
where~$I$ is the maximum of all of the arities~$I_w$ of all of the words,
instead of a quadratic form as in Eqs.~\ref{eq:tracker-relax},
\ref{eq:map-relax}, and~\ref{eq:event-relax}.

Eqs.~\ref{eq:sentence-relax}, \ref{eq:tracker-relax}, \ref{eq:map-relax},
and~\ref{eq:event-relax}, with unknowns taking binary values, are binary
integer-programming problems (linearly constrained in our case) and are
difficult to optimize with a large number of unknowns.
Thus we solve relaxed variants of these problems, taking the domain of the
indicator variables to be $[0,1]$ instead of $\{0,1\}$.
With the unchanged sum-to-one constraints, the subcollections~$p^{t,l}$
and~$q^{t,w}$ of the indicator variables~$p^{t,l}_j$ and~$q^{t,w}_k$ over
all~$j$ and~$k$, respectively, can be viewed as discrete distributions.
We show in Section~\ref{sec:relaxation} that, despite such relaxation,
Eqs.~\ref{eq:tracker-relax}, \ref{eq:map-relax}, \ref{eq:event-relax},
and~\ref{eq:sentence-relax} have the same global optima as
Eqs.~\ref{eq:tracker}, \ref{eq:map}, \ref{eq:event}, and~\ref{eq:sentence},
respectively.
Note that Eqs.~\ref{eq:tracker-relax}, \ref{eq:map-relax},
and~\ref{eq:event-relax} are all linearly constrained quadratic-programming
problems.
In the typical use case of the sentence tracker, where words have maximal arity
two (\ie\ the words in the sentence describe only pairwise relations between
nouns), Eq.~\ref{eq:sentence-relax} becomes linearly constrained cubic
programming.

\section{Relaxation and optimization}
\label{sec:relaxation}

An important property of this formulation is that the global optima of
Eq.~\ref{eq:sentence-relax} (and thus Eqs.~\ref{eq:tracker-relax},
\ref{eq:map-relax}, and~\ref{eq:event-relax} as special cases) are the same
whether the domains of the indicator variables are discrete or continuous,
\ie\ $\{0,1\}$ or $[0,1]$.
We prove this below.
This allows us to introduce a local-search technique to solve the relaxed
problems.

Eq.~\ref{eq:sentence-relax} has the following general form:
\begin{equation}
  \hspace*{-5pt}\scalebox{0.94}
  {\begin{array}[t]{l}
    \displaystyle\max_{\mathbf{x}}
    \sum_{n=1}^N\hspace*{-3pt}\left(\hspace*{-3pt}
    \mathop{\sum_{v_{u_1}\in V_{u_1}}\hspace*{-7pt}\cdots\hspace*{-7pt}\sum_{v_{u_n}\in V_{u_n}}}
    \limits_{\{u_1,\ldots,u_n\}\in\mathcal{E}_n}
    \hspace*{-8pt}\Phi(v_{u_1},\ldots,v_{u_n})
    x^{u_1}_{v_{u_1}}\cdots x^{u_n}_{v_{u_n}}
    \hspace*{-5pt}\right)\\
    \begin{aligned}
      \st\ &(\forall u)(\forall v_u)x_{v_u}^u\ge0\\
           &(\forall u)\sum_{v_u\in V_u} x^u_{v_u}=1\\
    \end{aligned}\\
  \end{array}}
  \label{eq:score-general}
\end{equation}
where~$x^u_{v_u}$ are (indicator) variables that form a polynomial objective,
$\Phi(v_{u_1},\ldots,v_{u_n})$ are the coefficients of the terms in this
polynomial, and we sum over all terms of the varying degree~$n\le N$.

This polynomial can be viewed as denoting the overall compatibility of a
labeling of an undirected hypergraph with vertices~$u$, labeled~$v_u$ from a
set~$V_u$ of labels, where each term denotes the compatibility of a possible
labeling configuration for the vertices of some hyperedge.
$\mathcal{E}_n$ denotes the set of all hyperedges $\{u_1,\ldots,u_n\}$ of
size~$n$.
The inner nested summation over the indices~$v_{u_1},\ldots,v_{u_n}$ sums over
all possible labeling configurations for a particular hyperedge.
Summing this over all elements of $\mathcal{E}_n$ sums over all hyperedges of
size~$n$.
The overall polynomial sums over all hyperedges of size $n\le N$.
In the limiting case of maximal arity two, the hypergraph becomes an ordinary
undirected graph and the hyperedges become ordinary undirected edges.

The collection~$x^u$ of variables $x^u_{v_u}$ can be viewed as a discrete
distribution over possible labels~$v_u\in V_u$ for vertex~$u$.
The coefficient $\Phi(v_{u_1},\ldots,v_{u_n})$ associated with each hyperedge
measures the compatibility of the labels for the constituent vertices.

This hypergraph can be viewed as a high-order Markov Random Field (MRF).
The common case of an MRF is when $N\le2$.
Ravikumar and Lafferty~\cite{Ravikumar2006} showed that, for this case, the
global optima of Eq.~\ref{eq:score-general} are the same whether the domains of
the variables are discrete or continuous, \ie\ $\{0,1\}$ or $[0,1]$.
Below, we generalize this result to $N\ge1$.

\begin{theorem}
  The optimal value of Eq.\ref{eq:score-general} is the same irrespective of
  whether the domains of the variables are $\{0,1\}$ or $[0,1]$.
\end{theorem}

\emph{Proof.} Let the optimal value before relaxation be $S_{\text{INT}}$ and
the optimal value after relaxation be~$S^*$.
It is obvious that $S^*\ge S_{\text{INT}}$ since $\{0,1\}\subseteq[0,1]$.
Now we show that $S_{\text{INT}}\ge S^*$.
Let $\mathbf{x}^*$ be the solution to the relaxed problem.
Under a probabilistic interpretation, the inner nested summation over the
indices~$v_u$ constitutes an expectation of the hyperedge compatibility.
Thus for a particular set~$\mathbf{x}^*$ of distributions, there must be some
labeling~$\overline{\mathbf{x}}$ with compatibility
$S(\overline{\mathbf{x}})\ge S^*$.
Since $S_{\text{INT}}$ is the value of the optimal labeling,
$S_{\text{INT}}\ge S(\overline{\mathbf{x}})$.
Thus $S_{\text{INT}}\ge S^*$. $\square$

Given a solution to the relaxed problem, we can obtain a discrete solution to
the original problem with the following algorithm that performs a single pass
over the vertices:
\begin{compactenum}
\item Label any vertex~$u$ for which $x^u_{v_u}\in\{0,1\}$, with the
  label~$v_u^*$ for which $x^u_{v_u^*}=1$.
\item Label any unlabeled vertex~$u$ which does not satisfy the above condition
  with the label~$v_u^*$ that maximizes Eq.~\ref{eq:score-general} while
  keeping the label distributions of other vertices unchanged.
  Then set $x^u_{v_u^*}=1$ and $x^u_{v_u}=0$ if $v_u\neq v_u^*$.
\end{compactenum}
Repeat this process until all vertices are labeled.
Since at each step the overall compatibility does not decrease, the resulting
discrete labeling will yield overall compatibility at least as good as that of
the relaxed problem.
However, since the relaxed overall compatibility is optimal, the discrete
solution constructed by the above process must be equivalent to the continuous
one.
Thus a global optimum of the relaxed continuous optimization problem is also a
global optimum of the original combinatorial optimization problem.

Unfortunately, there is no effective method for finding the global optima of
nonlinear objectives such as the relaxed objective in
Eq.~\ref{eq:score-general}.
Local search methods such as gradient descent are often used to find local
optima of constrained nonlinear programs.
While local search is not guaranteed to find a global optimum, we demonstrate
empirically in Section~\ref{sec:experiments} that local search generates
solutions qualitatively identical to the global optimum for
Eq.~\ref{eq:sentence-relax}.

The discrete formulation of Eqs.~\ref{eq:tracker}, \ref{eq:map},
and~\ref{eq:event}, constitutes combinatorial optimization.
Such is feasible with the Viterbi algorithm because it exploits the sequential
structure of these objectives, allowing it to find a global optimum in
polynomial time.
More specifically, the graph structure is layered in the form of a lattice.
The Viterbi algorithm enumerates all possible labeling configurations at each
layer in this lattice.
The Viterbi algorithm not only explicitly enumerates all possible labeling
configurations, it achieves optimality by saving all such at one layer in the
lattice to search through while computing the next layer.
Thus both the space and time complexity of the Viterbi algorithm depends on
enumeration of all possible labeling configurations.
This number of possible labeling configurations is small for
Eqs.~\ref{eq:tracker}, \ref{eq:map}, and~\ref{eq:event} but becomes exponential
in~$L$ and~$W$ for Eq.~\ref{eq:sentence}.

In contrast, the relaxation approach never explicitly enumerates all possible
labeling configurations; such enumeration is performed implicitly by local
search.
Moreover, it never stores such.
Since there is no explicit enumeration in the relaxation approach, the space
complexity does not depend on the number of possible labeling configurations
and is polynomial in~$T$, $J$, $K$, $L$, and~$W$.
For Eq.~\ref{eq:sentence-relax}, the maximal hyperedge size is the maximal
arity of the semantic primitives used to represent the meanings of words out of
which the meanings of sentences are constructed.
This will always be independent of, and much smaller than, the sentence length.
If the maximal hyperedge size is constant, then the space complexity of
Eq.~\ref{eq:sentence-relax} is polynomial.
Furthermore, in most cases the maximal arity is two and thus the maximal
hyperedge size is three, which limits the polynomial degree to cubic objectives.

We know of no analytical bound on the time complexity of the relaxation
approach; however one can construct, as we do below, a local search method
where each iteration takes polynomial time but we know of no bound on the
number of iterations.
Intuitively, one can view the iterative process as searching through
possible labeling configurations.
Performing local search in the continuous domain can avail itself of gradient
information that is unavailable when doing discrete combinatorial optimization.
%
% yu239: BP, trw-s, and GBP are for combinatorial optimizations
% maybe we should discuss them before the relaxation
%
Below we show how to use efficient local search to find a local optimum of the
objective in Eq.~\ref{eq:score-general}.
Note that the objective may have degree greater than two.
Thus we cannot employ conventional methods (\eg\ belief propagation
\cite{Pearl1982,Pearl1988} and tree-reweighted message passing
\cite{Kolmogorov2006}) that solve MRFs which correspond to polynomial
objectives of degree two.
We instead employ the Growth Transform (GT)~\cite{Baum1967} to solve this
constrained nonlinear programming problem as it applies to polynomials of any
degree.
Other methods, such as Generalized Belief Propagation (GBP)~\cite{Yedidia2000},
that apply to hyperedges of any order, may be used as well.

GT is a local-search technique for optimizing polynomial objectives.
It is a generalized form of the Baum-Welch reestimation
procedure~\cite{baum1972,baumpsw1970} which is widely used to perform
maximum-likelihood estimation of HMM parameters.
In order to apply GT, the objective must satisfy three conditions:
\begin{compactenum}[(a)]
\item The objective $P(\mathbf{x})=P(\{x^u_{v_u}\})$ must be a
  homogeneous polynomial, \ie\ all terms must have the same degree.
  \label{a}
\item The coefficients must be nonnegative.
  \label{b}
\item The variables~$x^u_{v_u}$ must be nonnegative and satisfy the
  sum-to-one constraint: $\sum_{v_u}x^u_{v_u}=1$ for each~$u$.
  \label{c}
\end{compactenum}

GT maximizes the objective iteratively with the following update formula:
\begin{equation}
  x^u_{v_u}\leftarrow
  Z^ux^u_{v_u}
  {\frac{\partial P}{\partial x^u_{v_u}}}\bigg|_{\mathbf{x}}
  \label{eq:update}
\end{equation}
where~$Z^u$ is an iteration-dependent normalization factor to maintain the
sum-to-one constraint.
We now show that Eq.~\ref{eq:sentence-relax}, as formulated as a special case of
Eq.~\ref{eq:score-general}, can be modified to satisfy these conditions without
changing the objective.
Clearly, condition~(\ref{c}) is already met.
Conditions~(\ref{a}) and~(\ref{b}) can be satisfied by leveraging
condition~(\ref{c}).
Suppose the maximal degree of the objective is~$N$.
Any term with degree $m<N$ can be converted to a form with degree
$N$ by adding redundant free variables as follows:
\begin{equation}
  \hspace*{-15pt}\scalebox{0.75}
  {\begin{array}[t]{@{}l@{}}
      \displaystyle
    \sum_{v_{u_1},\ldots,v_{u_m}}
    \Phi(v_{u_1},\ldots,v_{u_m})
    x^{u_1}_{v_{u_1}}\cdots x^{u_m}_{v_{u_m}}\\
      \displaystyle
    {}=\sum_{v_{u_1},\ldots,v_{u_m}}
    \left(\sum_{v_{u_{m+1}},\ldots,v_{u_n}}
    x^{u_{m+1}}_{v_{u_{m+1}}}\cdots x^{u_n}_{v_{u_n}}
    \Phi(v_{u_1},\ldots,v_{u_m})\right)
    x^{u_1}_{v_{u_1}}\cdots x^{u_m}_{v_{u_m}}\\
      \displaystyle
    {}=\sum_{v_{u_1},\ldots,v_{u_m}}
    \sum_{v_{u_{m+1}},\ldots,v_{u_n}}
    \Phi(v_{u_1},\ldots,v_{u_m})
    x^{u_1}_{v_{u_1}}\cdots x^{u_m}_{v_{u_m}}
    x^{u_{m+1}}_{v_{u_{m+1}}}\cdots x^{u_n}_{v_{u_n}}
  \end{array}}\hspace*{-30pt}
\end{equation}
The objective can be made homogeneous by applying the above transformation to
each hyperedge of the original nonhomogeneous objective, thus satisfying
condition~(\ref{a}).
This transformation does not change the gradient with respect to existing
variables, and thus does not effect the GT update formula.
Therefore its mere existence demonstrates that GT can apply even without
performing the transformation.
This is important because the transformation would increase the space and time
complexity of GT.\@

To satisfy condition~(\ref{b}), we increase each coefficient in the objective
with a term-dependent positive constant:
\begin{equation}
  \scalebox{0.9}
  {\begin{array}[t]{@{}l@{}}
    C_{\{u_1,\ldots,u_m\}}=
    \max(\epsilon,-\min_{v_{u_1},\ldots,v_{u_m}}\Phi(v_{u_1},\ldots,v_{u_m}))
  \end{array}}
  \label{eq:constant}
\end{equation}
where $\epsilon>0$.
We add each such constant to its corresponding term:
\begin{equation}
  \hspace*{-20pt}
  \scalebox{0.9}
  {\begin{array}[t]{l}
      \displaystyle
      \sum_{v_{u_1},\ldots,v_{u_m}}
      \left[\Phi(v_{u_1},\ldots,v_{u_m}) + C_{\{u_1,\ldots,u_m\}}\right]
      x^{u_1}_{v_{u_1}}\cdots x^{u_M}_{v_{u_m}}\\
      \displaystyle
    {}=\begin{array}[t]{l}
    \displaystyle
    \sum_{v_{u_1},\ldots,v_{u_m}}
      \Phi(v_{u_1},\ldots,v_{u_m})
      x^{u_1}_{v_{u_1}}\cdots x^{u_m}_{v_{u_m}}\\
    \displaystyle
      {}+\sum_{v_{u_1},\ldots,v_{u_m}}
      x^{u_1}_{v_{u_1}}\ldots x^{u_m}_{v_{u_m}}
      C_{\{u_1,\ldots,u_m\}}
      \end{array}\\
    \displaystyle
    {}=C_{\{u_1,\ldots,u_m\}}+\sum_{v_{u_1},\ldots,v_{u_m}}
    \Phi(v_{u_1},\ldots,v_{u_m})
    x^{u_1}_{v_{u_1}}\ldots x^{u_m}_{v_{u_m}}\\
  \end{array}}\hspace*{-20pt}
\end{equation}
This term now has positive coefficients, and thus so does the whole objective,
yet, the objective is unchanged.

When applying GT to Eqs.~\ref{eq:sentence-relax}, \ref{eq:tracker-relax},
\ref{eq:map-relax}, and~\ref{eq:event-relax}, one must store the indicator
variables.
However, one need not store the coefficients; one could recompute them at every
iteration.
Thus the space complexity is driven by the indicator variables:
$O(TJ)$ for Eq.~\ref{eq:tracker-relax},
$O(TK)$ for Eq.~\ref{eq:map-relax},
$O(T(J+K))$ for Eq.~\ref{eq:event-relax}, and
$O(T(JL+KW))$ for Eq.~\ref{eq:sentence-relax}.
Each iteration of Eq.~\ref{eq:update} is dominated by the time to compute the
gradient.
Doing so with reverse-mode automatic differentiation (AD)
\cite{Bryson-1962a,Bryson-Ho-1969a,Griewank-2012a,Ostrovskii1971UdB,Speelpenning80,Werbos-1974a}
takes the same time as computing the objective.
Thus the time complexity of each iteration is:
$O(TJ^2)$ for Eq.~\ref{eq:tracker-relax},
$O(TK^2)$ for Eq.~\ref{eq:map-relax},
$O(T(J+K)^2)$ for Eq.~\ref{eq:event-relax}, and
$O(T(LJ^2+W(J^IK+K^2)))$ for Eq.~\ref{eq:sentence-relax}.

\section{Experiments}
\label{sec:experiments}

We have shown so far that
\begin{compactitem}
\item Eqs.~\ref{eq:tracker-relax}, \ref{eq:map-relax}, \ref{eq:event-relax},
  and \ref{eq:sentence-relax} yield the same optima as Eqs.~\ref{eq:tracker},
  \ref{eq:map}, \ref{eq:event}, and \ref{eq:sentence} when the indicator
  variables are constrained to be in $\{0,1\}$ and
\item The global optima for Eqs.~\ref{eq:tracker-relax}, \ref{eq:map-relax},
  \ref{eq:event-relax}, and \ref{eq:sentence-relax} are the same
  whether the domains of the indicator variables are discrete or continuous,
  \ie\ $\{0,1\}$ or $[0,1]$.
\end{compactitem}
Moreover, we have presented a GT method for performing local search for local
optima of Eqs.~\ref{eq:tracker-relax}, \ref{eq:map-relax},
\ref{eq:event-relax}, and \ref{eq:sentence-relax}.
We now present empirical evidence that the local optima produced by GT are
qualitatively identical to the global optima.

When performing GT, we employed the following procedure.
We randomly initialized the local search at 150 different label distributions.
For each one, we ran 300 iterations of Eq.~\ref{eq:update}.
We then selected the resulting label distribution that corresponded to the
highest objective and ran 5000 additional iterations on this label distribution
to yield the resulting label distribution and objective.
Performing 150 restarts may only be necessary for problems with a large number
of variables.
Nonetheless, for simplicity, we employed this uniform number of restarts for
all experiments.

\begin{figure}
  \centering
  \resizebox{\columnwidth}{!}{\begin{tabular}{@{}cc@{}}
    \includegraphics[width=0.5\columnwidth]{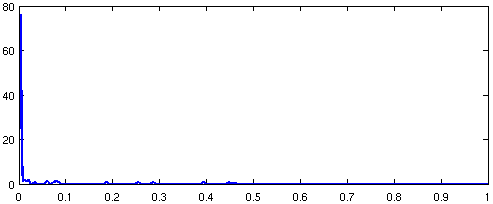}
    &\includegraphics[width=0.5\columnwidth]{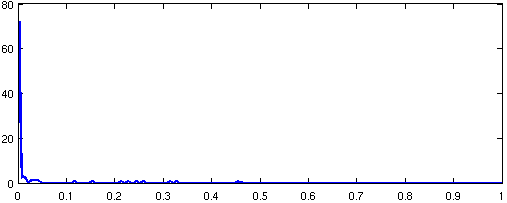}\\
    $J=30,K=3$&$J=120,K=3$\\
    \includegraphics[width=0.5\columnwidth]{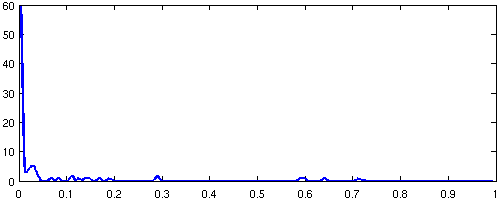}
    &\raisebox{20pt}{intractable for Viterbi}\\
    $J=30,K=9$&$J=120,K=9$\\
  \end{tabular}}
  \caption{Histograms of the number of videos with given ranges of relative
    error.}
  \vspace*{-2ex}
  \label{fig:histogram}
\end{figure}

We evaluated the GT method on the same dataset that was used by Yu and Siskind
\cite{Yu2013}\footnote{\url{http://upplysingaoflun.ecn.purdue.edu/~qobi/acl2013-dataset.tgz}}.
This dataset contains 94 video clips, each with between two and five sentential
annotations of activities that occur in the corresponding clip.
All but one of these clips contain at least one annotated sentence with exactly
three content words: a subject, a verb, and a direct object.
We did not process the single clip that does not.
For the remaining 93 video clips, we selected the first annotated sentence with
exactly three content words.
For each such video-sentence pair, we computed the global optimum to
Eq.~\ref{eq:sentence} with the Viterbi algorithm using the same features and
hand-crafted word HMMs (for $K=3$, for $K=9$ extra states were added) as used
by Yu and Siskind
\cite{Yu2013}\footnote{\url{http://github.com/yu239/sentence-training}}.
For each video-sentence pair, we also computed a local optimum to
Eq.~\ref{eq:sentence-relax} with GT using these same features and HMMs.
We then computed relative error (percentage) between the global optimum
computed by the Viterbi algorithm and local optimum computed by GT for each
video-sentence pair, and averaged over all 93 pairs.
We repeated this three times:
\begin{center}
\vspace*{-2ex}
\resizebox{\columnwidth}{!}{
  \begin{tabular}{@{}c|c|c@{}}
    &$J=30$&$J=120$\\
    \hline
    $K=3$
    &\makecell[l]
        {five top-scoring detections\\
          for each of six object classes\\
          and HMMs with up to three states}
    &\makecell[l]
        {twenty top-scoring detections\\
          for each of six object classes\\
          and HMMs with up to three states}\\
    \hline
    $K=9$
    &\makecell[l]
        {five top-scoring detections\\
          for each of six object classes\\
          and HMMs with up to nine states}
    &\makecell[l]
        {can run with GT, but Viterbi, needed\\
          for global optimum to compute\\
          relative error, is intractable}\\
  \end{tabular}}
\end{center}
The average relative errors for $(J=30,K=3)$, $(J=120,K=3)$, and $(J=30,K=9)$,
are 2.22\%, 2.79\%, and 5.08\%, respectively.
Fig.~\ref{fig:histogram} gives histograms of the number of videos with given
ranges of relative error.

\begin{figure*}
  \centering
  \begin{tabular}{@{}cccccc@{}}
    \includegraphics[width=0.16\textwidth]{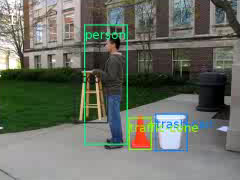}&
    \includegraphics[width=0.16\textwidth]{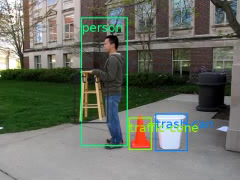}&
    \includegraphics[width=0.16\textwidth]{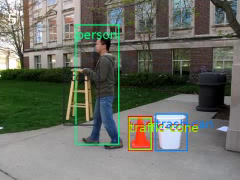}&
    \includegraphics[width=0.16\textwidth]{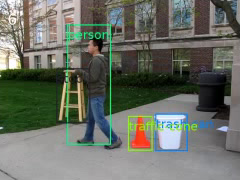}&
    \includegraphics[width=0.16\textwidth]{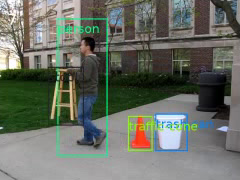}&
    \includegraphics[width=0.16\textwidth]{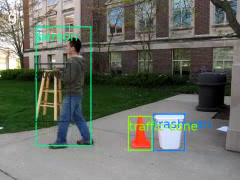}\\
    \multicolumn{6}{l}{\emph{The person to the left of the traffic-cone carried
        the stool to the left of the trash-can slowly away from the
        traffic-cone.}}\\
    \multicolumn{6}{l}{56s, $T=12$, $I=2$, $J=30$, $K=3$, $L=5$, $W=10$}\\
    \includegraphics[width=0.16\textwidth]{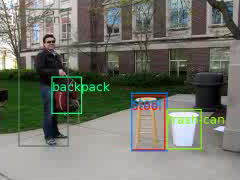}&
    \includegraphics[width=0.16\textwidth]{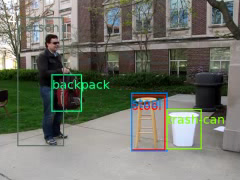}&
    \includegraphics[width=0.16\textwidth]{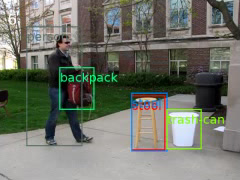}&
    \includegraphics[width=0.16\textwidth]{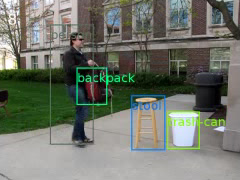}&
    \includegraphics[width=0.16\textwidth]{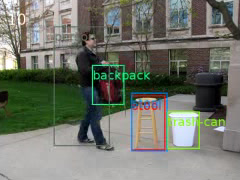}&
    \includegraphics[width=0.16\textwidth]{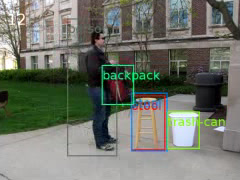}\\
    \multicolumn{6}{l}{\emph{The person to the left of the trash-can carried
        the backpack to the left of the stool slowly towards the stool.}}\\
    \multicolumn{6}{l}{61s, $T=15$, $I=2$, $J=30$, $K=3$, $L=5$, $W=10$}\\
    \includegraphics[width=0.16\textwidth]{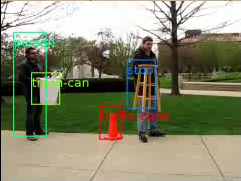}&
    \includegraphics[width=0.16\textwidth]{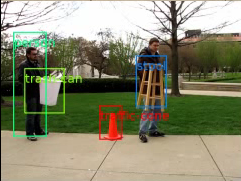}&
    \includegraphics[width=0.16\textwidth]{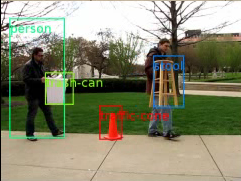}&
    \includegraphics[width=0.16\textwidth]{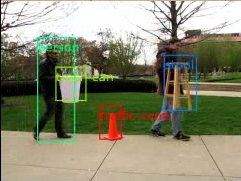}&
    \includegraphics[width=0.16\textwidth]{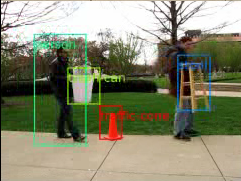}&
    \includegraphics[width=0.16\textwidth]{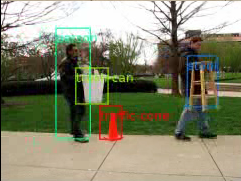}\\
    \multicolumn{6}{l}{\emph{The person to the left of the stool carried the trash-can
slowly towards the traffic-cone.}}\\
    \multicolumn{6}{l}{129s, $T=15$, $I=2$, $J=30$, $K=9$, $L=4$, $W=8$}\\
    \includegraphics[width=0.16\textwidth]{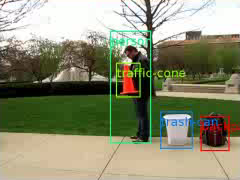}&
    \includegraphics[width=0.16\textwidth]{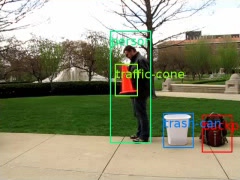}&
    \includegraphics[width=0.16\textwidth]{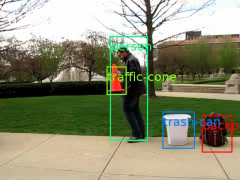}&
    \includegraphics[width=0.16\textwidth]{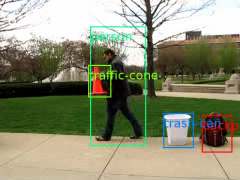}&
    \includegraphics[width=0.16\textwidth]{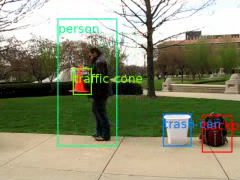}&
    \includegraphics[width=0.16\textwidth]{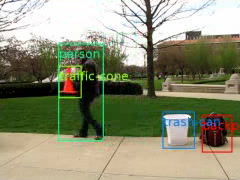}\\
    \multicolumn{6}{l}{\emph{The person to the left of the trash-can carried
        the traffic-cone to the left of the backpack.}}\\
    \multicolumn{6}{l}{667s, $T=15$, $I=2$, $J=120$, $K=3$, $L=4$, $W=7$}\\
    \includegraphics[width=0.16\textwidth]{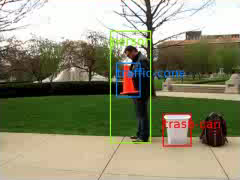}&
    \includegraphics[width=0.16\textwidth]{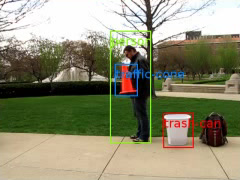}&
    \includegraphics[width=0.16\textwidth]{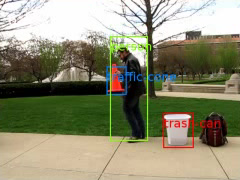}&
    \includegraphics[width=0.16\textwidth]{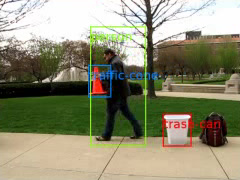}&
    \includegraphics[width=0.16\textwidth]{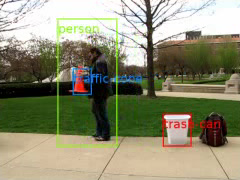}&
    \includegraphics[width=0.16\textwidth]{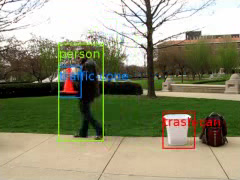}\\
    \multicolumn{6}{l}{\emph{The person to the left of the trash-can carried
        the traffic-cone.}}\\
    \multicolumn{6}{l}{520s, $T=15$, $I=2$, $J=120$, $K=3$, $L=3$, $W=5$}\\
    \includegraphics[width=0.16\textwidth]{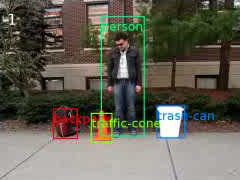}&
    \includegraphics[width=0.16\textwidth]{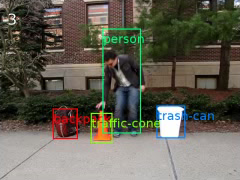}&
    \includegraphics[width=0.16\textwidth]{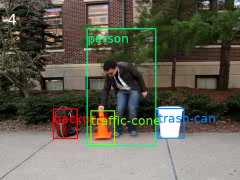}&
    \includegraphics[width=0.16\textwidth]{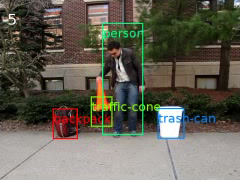}&
    \includegraphics[width=0.16\textwidth]{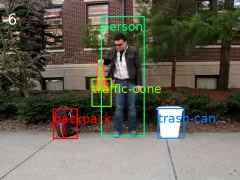}&
    \includegraphics[width=0.16\textwidth]{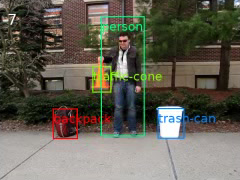}\\
    \multicolumn{6}{l}{\emph{The person to the left of the trash-can picked up
        the traffic-cone to the right of the backpack quickly.}}\\
    \multicolumn{6}{l}{230s, $T=10$, $I=2$, $J=90$, $K=3$, $L=4$, $W=8$}\\
  \end{tabular}
  \caption{Tracks produced by processing six video clips of varying lengths~$T$
    with the associated sentences of varying complexity~$L$ and~$W$, along with
    the running time in seconds to produce such with GT from the indicated
    number of detections~$J$ per frame and word models of varying
    complexity~$I$ and~$K$.
    Best viewed on a color screen.}
  \vspace*{-2ex}
  \label{fig:videos}
\end{figure*}

We limited the above experiment to sentences with no more than three content
words because we wished to compare against the global optimum which can only be
computed with the Viterbi algorithm.
This process may become intractable with sentences with more than three content
words, especially with large~$J$.
To demonstrate that the local-search approach can scale to process longer
sentences we performed a second experiment.
In this experiment, we processed six video clips with longer sentences.
Fig.~\ref{fig:videos} illustrates the tracks produced for these video-sentence
pairs, as derived from the indicator variables~$p^{t,l}_j$.
Note that for the example in the fourth row in Fig.~\ref{fig:videos}, computing
the global optimum with the Viterbi algorithm would take
$15\times(120^4\times3^1)^2=5,804,752,896,000,000,000$ lattice comparisons.
We estimate that this would take about 20 years on a current computer.
%
% an example of the space difference between viterbi and relaxation form
% relaxation: O(T*(W*K+L)*J*J) (if only pair-wise features)
% viterbi: T*J^L*K^W
% if J=30,T=20,L=4,K=3,W=3, and double array (8bytes)
% relaxation: 20*3*3*30*30 = 162000 => 1MB
% viterbi: 30*30*30*30*3*3*3*20=437400000 => 3337MB(3.3G)
%
\setlength{\tabcolsep}{1pt}

\section{Related work}

Lin \etal\ \cite{Lin2014} present an approach for searching a database of
video clips for those that match a sentential query.
Implicit in this is a function that scores how well a video clip depicts a
sentence or alternatively how well a sentence describes a video clip.
That work differs from the current work in several ways.
First, they run a tracker in isolation, independently for each object, before
scoring a video-sentence pair.
Here, we jointly perform tracking and scoring, and do so jointly for all tracks
described by a sentence.
This allows scoring to influence tracking and tracking one object to influence
tracking other objects.
Second, their scoring function composes only unary primitive predicates, each
applied to a single track.
Here, our scoring function composes multivariate primitive predicates, each
applied to multiple tracks.
This is what links the word meanings together and allows the entire sentential
semantics to influence the joint tracking of all participants.
Third, the essence of their recognition process (not training) is that they
automatically find what they denote as~$y_{uv}$.
This maps tracks~$v$ to (unary) arguments of primitive predicates~$u$.
Here, we solve a dual problem of automatically finding what we denote
as~$p^l_j$, a map from event participants~$l$ to detections~$j$.
Fourth, they allow predicates to be assigned to a dummy track ``no-obj'' which
allows them to ignore portions of the sentence.
Here, we constrain the track collection to satisfy all primitive predicates
contained in a sentence.
Fifth, they adopt a ``no coreference'' constraint that requires that each
nondummy track be assigned to a different primitive predicate.
Here, we allow such.
This allows us to process sentences like \emph{The person approached the
  backpack to the left of the chair} that contain, in part, primitive
predicates like
$\textsc{approach}(\textbf{person},\textbf{backpack})\wedge\textsc{leftOf}(\textbf{backpack},\textbf{chair})$
where the track $\textbf{backpack}$ is assigned to two different primitive
predicates.
Sixth, they do not model temporally-varying features in their primitive
predicates.
Here, we do so with HMMs.

The relaxed sentence tracker also shares some similarity in spirit with current
research in image/video object co-detection/discovery
\cite{Hayder2014,Joulin2014,Prest2012,Tang2014} based on object proposals
\cite{Alexe2012,Cheng2014,Zitnick2014}.
The goal of this line of work is to find instances of a common object across a
set of different images or video frames, given a pool of object candidates
generated by measuring the `objectness' of image windows in each image.
This is done by associating a unary cost with each candidate to represent the
confidence that that candidate is an object, and a binary cost between pairs of
candidates to measure their similarity in appearance, resulting in a
second-order MRF/CRF.
Selecting the best candidate for each image constitutes a MAP estimate on this
random field, and is usually relaxed to constrained nonlinear programming
\cite{Joulin2014,Tang2014}, or computed by a combinatorial optimization
technique such as tree-reweighted message passing
\cite{Kolmogorov2006,Prest2012}.
Except for its loopy graph structure, this formulation is analogous to
detection-based tracking in Eq.~\ref{eq:tracker-relax}.
In contrast to the work presented here, prior work on object
co-detection/discovery only exploits visual appearance---analogous to the
term~$f(b)$ in Eq.~\ref{eq:tracker-relax}---and object-pair
similarity---analogous to the term $g(b',b)$ in
Eq.~\ref{eq:tracker-relax}---but not the degree to which an object track
exhibits particular spatiotemporal behavior, as is done by the terms $h(k,b)$
and $a(k',k)$ in the event tracker, Eq.~\ref{eq:event-relax}, let alone the
joint detection/discovery of multiple objects that exhibit the collective
spatiotemporal behavior described by a sentence, as is done in the sentence
tracker, Eq.~\ref{eq:sentence-relax}.

\section{Discussion}

Siddharth \etal\ \cite{siddharth2014}, Barbu \etal \cite{barbu2014}, and
Yu and Siskind \cite{Yu2013} present a variety of applications for the sentence
tracker.
Since the sentence tracker is simply a scoring function that scores a
video-sentence pair, one can use it for the following applications:
\begin{compactdesc}
\item[focus of attention] Given a single video clip as input, that contains two
  different activities taking place with different subsets of participants,
  along with two different sentences that each describe these different
  activities, produce two different track collections that delineate the
  participants in the two different activities.
\item[sentence generation] Given a video clip as input, produce a sentence as
  output that best describes the video clip by searching the space of all
  possible sentences to find the one with the highest score on the input video
  clip.
\item[video retrieval] Given a sentential query as input, search a dataset of
  video clips to find that clip that best depicts the query by searching all
  clips to find the one with the highest score on the input query.
\item[language acquisition] Given a training corpus of video clips paired with
  sentences that described these clips, search the parameter space of the
  models for the words in the lexicon that yields the highest aggregate score
  on the training corpus.
\end{compactdesc}
With the exception of the last use case, language acquisition, the other use
cases all treat the sentence tracker as a black box.
We have demonstrated a plug-compatible black box that takes the same input in
the form of a video-sentence pair and produces the same output in the form of a
score and a track collection.
Since we have proven that the global optimum to the relaxed objective is the
same as that for the original objective, and further empirically demonstrated
that local search tends to find a local optimum that is qualitatively the same
as the global one, one can employ this relaxed method to perform exactly the
same first three uses cases as the original sentence tracker with identical, or
nearly identical results.
The chief advantage of the relaxed method is that it scales to longer sentences.
More precisely, we demonstrate three distinct kinds of scaling:
\begin{compactenum}
\item Because the original sentence tracker had space complexity of $O(TJ^LK^W)$
  and time complexity $O(T(J^LK^W)^2)$ it was, in practice, limited to
  small~$J$, typically at most~20.
  Here, we perform experiments that demonstrate scaling to $J=120$.
\item Similarly, the original sentence tracker was limited, in
  practice, to small~$K$, typically at most~3.
  Here, we perform experiments that demonstrate scaling
  to $K=9$.\footnote{The upper bound of~$K$ is, in fact, determined by the
    number of frames, which is 15, on average, in our experiments}
\item Similarly, the original sentence tracker was limited, in practice, to
  small~$L$ and~$W$, typically $L\leq 3$, \ie\ at most three nouns in the
  sentence, and typically at most two words that have more than one state in
  their HMM.\@
  (Words that describe static properties, like nouns, adjectives, and
  spatial-relation prepositions, typically have a single-state HMM while words
  that describe dynamic properties, like verbs, adverbs, and motion
  prepositions, typically have multiple states.)
  Here, we perform experiments that demonstrate scaling to $L=5$ and
  $W=10$.
\end{compactenum}
Since the method of Yu and Siskind \cite{Yu2013} for the fourth use case,
namely language acquisition, does not use the sentence tracker as a black box,
extending our new method to this use case is beyond the scope of this current
work.

\section{Conclusion}

The sentence tracker has previously been demonstrated to be a powerful
framework for a variety of applications, both theoretical and potentially
practical.
It can serve as a model of grounded child language acquisition \cite{Yu2013} as
well as the basis for searching full-length Hollywood movies for clips that
match rich sentential queries \cite{barbu2014} in a way that is sensitive to
subtle semantic distinctions in the queries.
However, until now, it was impractical to apply in many situations that require
scaling to complex sentences with many event participants.
Our results remove that barrier.

\section*{Acknowledgments}
This research was sponsored, in part, by the Army Research Laboratory and was
accomplished under Cooperative Agreement Number W911NF-10-2-0060.
The views and conclusions contained in this document are those of the authors
and should not be interpreted as representing the official policies, either
express or implied, of the Army Research Laboratory or the U.S. Government.
The U.S. Government is authorized to reproduce and distribute reprints for
Government purposes, notwithstanding any copyright notation herein.

\end{document}